\documentclass[letterpaper, 10 pt, conference]{ieeeconf}  %
\pdfoutput=1

\IEEEoverridecommandlockouts                              %
\makeatletter
\let\NAT@parse\undefined
\makeatother

\usepackage[usenames, dvipsnames]{xcolor}
\usepackage[caption=false]{subfig}
\usepackage[]{graphicx} %
\usepackage[pdfborderstyle={/S/U/W 1}]{hyperref}
\usepackage{units}
\usepackage{booktabs}
\usepackage{paralist}
\usepackage[capitalize]{cleveref}

\colorlet{mylinkcolor}{BrickRed}
\colorlet{mycitecolor}{Green}
\colorlet{myurlcolor}{NavyBlue}

\hypersetup{
  linkcolor  = mylinkcolor,
  citecolor  = mycitecolor,
  urlcolor   = myurlcolor,
  colorlinks = true,
}

\usepackage[markup=bfit, deletedmarkup=sout, authormarkup=brackets]{changes}

\graphicspath{{./gfx/}}

\newcommand\T{\rule{0pt}{1.2ex}}       %
\newcommand\B{\rule[-1.2ex]{0pt}{0pt}} %

\title{\LARGE \bf
The HRC Model Set for Human-Robot Collaboration Research
}

\author{Sofya Zeylikman$^{1}$, Sarah Widder$^{2}$, Alessandro Roncone$^{2}$ , Olivier Mangin$^{2}$ and Brian Scassellati$^{2}$%
\thanks{ This work has been supported by a subcontract with Rensselaer Polytechnic Institute (RPI) by the Office of Naval Research, under Science and Technology: Apprentice Agents.}%
\thanks{$^{1}$ S. Z. is with the Center for Engineering, Innovation and Design, Yale University, New Haven, CT 06511, USA.}%
\thanks{$^{2}$ These authors are with the Social Robotics Lab, Computer Science Department, Yale University, New Haven, CT 06511, USA.}%
}

\begin{document}

\maketitle
\thispagestyle{empty}
\pagestyle{empty}

\begin{abstract}

In this paper, we present a model set for designing human-robot collaboration (HRC) experiments.
It targets a common scenario in HRC, which is the collaborative assembly of furniture, and it consists of a combination of standard components and custom designs.
With this work, we aim at reducing the amount of work required to set up and reproduce HRC experiments, and we provide a unified framework to facilitate the comparison and integration of contributions to the field.
The model set is designed to be modular, extendable, and easy to distribute.
Importantly, it covers the majority of relevant research in HRC, and it allows tuning of a number of experimental variables that are particularly valuable to the field.
Additionally, we provide a set of software libraries for perception, control and interaction, with the goal of encouraging other researchers to proactively contribute to our work.
 
\end{abstract}

\section{INTRODUCTION} %
\label{sec:introduction}

Robotics is a relatively recent experimental science. As with most scientific disciplines, it needs to provide guarantees for reproducibility and replicability.
However, as young as robotics is, it is co-evolving alongside the same enabling technologies that allow research in the field.
This can hamper the stability of the experimental tools and hence impedes the reproducibility of robotics experiments.
The problem is particularly relevant to human-robot collaboration (HRC), an important and growing area within robotics research.
HRC lacks standards for replicability, a factor that would accelerate progress and facilitate integration of advancements from independent contributions in the field.
This is primarily due to the fact that HRC targets a broad range of topics, from rescue robots to collaborative manufacturing and service robotics.
Defining what the term `\textsl{collaboration}' entails is actually still an open question and an integral part of the research in this domain.
Importantly, as of now, there exists no reference collaboration setup or standardized collaborative task in the field.
This paper explicitly targets this problem, with the goal of promoting ease of reproduction, evaluation, and integration of HRC research.

\begin{figure}
  \centering
  \includegraphics[width=.68\linewidth]{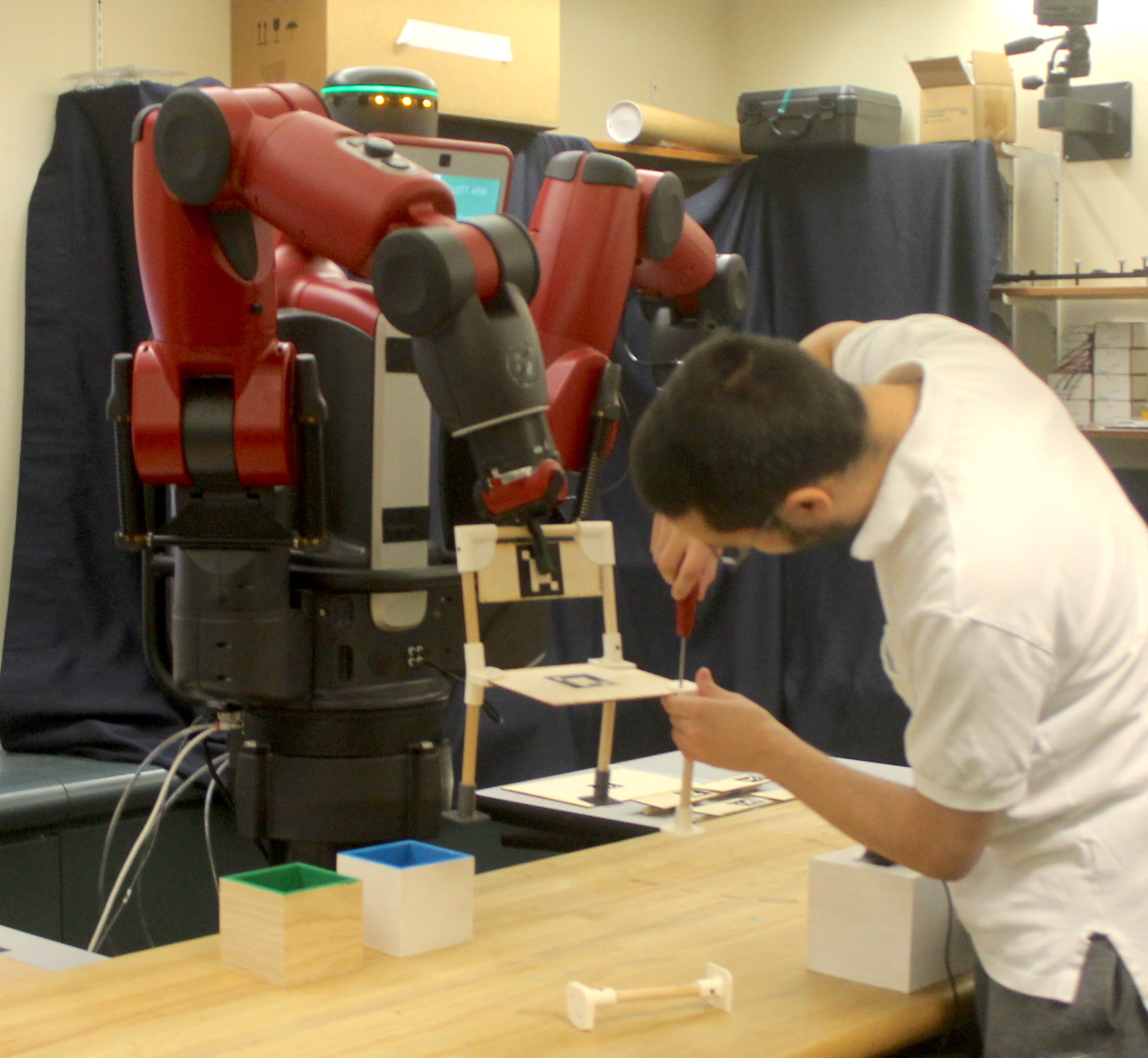}
  \caption{Human participant engaged in the joint construction of a chair with the Baxter Research Robot. More precisely, the robot is providing support by holding the chair in place while the human partner is screwing a leg.}\label{fig:setup}
\end{figure}

We identify a number of critical points related to the problem of replicability and standardization of research in HRC---most of which apply to the broader robotics field as well.
Firstly, research teams use a variety of robotic platforms that, for the most part, lack standardization.
Secondly, experiment design is affected by limitations in the enabling technologies, e.g. inadequate perception/manipulation skills.
Lastly, specific limitations of the robotic platform itself constrain the task domain.
For example, as highlighted in \cref{sec:background}, of the few works that actively use a robot to physically engage in a collaborative task, a common choice is to rely on flat-pack furniture \cite{Roncone2017,Knepper2013,Tellex2014a,Rozo2013}.
Whilst this solution is advantageous for a number of reasons (i.e. big, flat components that are easy to grasp combined with fine manipulation tasks to be performed by the human partner), in many cases the final designs exceed the payload of a standard collaborative robot, which often is lower than $\unit[5]{kg}$---e.g. the Baxter robot has a payload of $\unit[2.2]{kg}$.
As a result, the robot may not be physically capable of handling a completed (or close to completion) assembly, thus limiting its efficacy as a partner.
Other elements that are relevant to the design of reusable experiments are:
scalability to real-life problems,
(low) cost and (high) availability of the constituent parts,
ability to be intuitively understood by naive participants, and
flexibility in terms of cognitive skills required (e.g. assembling a complex electronic circuit vs stacking blocks).

Not only do these elements hinder the design and replication of HRC experiments, but they also obfuscate evaluation and comparison of research.
This is especially true when dissemination of the technology is deemed secondary to the scientific contribution.
To address the issue, separation between task domain and algorithmic contribution needs to be encouraged, in order for comparative studies to aim attention at the algorithm itself while `freezing` all the components related to robot platform and experimental evaluation.
Further, lack of standardization and reproducibility impedes efforts to integrate advancements from within the field itself and outside areas.
Hence, we claim that the absence of a unified \emph{language} for HRC experiments strongly limits the comparative evaluation of methods from the field.

In this article, we introduce a common ground for designing HRC tasks that is intended to be reused across a wide range of experiments.
To this end, we aim at comprehensively covering the broad spectrum of works in HRC, with the goal of devising a reference design for which it is possible to tune the many aspects of collaboration as parameters.
Specifically, we target the following properties:
\begin{inparaenum}[i)]
\item the robot is an effective contributor to the collaborative task---which rules out many domains due to current limitations of modern robotics systems;
\item the target domain enables by design a variety of capabilities (in terms of perception, control, and reasoning skills of both partners), to account for aspects of role assignment between peers---should it be predefined, negotiated, or take the form of one peer supporting the other;
\item the task is at least partially perceivable/understandable by both partners for them to share a common goal, and tuning information asymmetry in favor of either peer is possible;
\item communication---used to impart orders, negotiate, disclose internal states, or gather information---is treated as a first-class citizen in the design of the task;
\item the target domain is scalable in terms of complexity of the task and complexity of the human-robot interaction.
\end{inparaenum}

In order to improve permeability between independent research in the field, we propose a reference set of tasks for reuse in HRC experiments that aims at easing implementation and reproducibility, and allows adaptation to a variety of robot environments and capabilities.
We present a set of basic components that can be built into a variety of final designs, thus providing multiple collaborative tasks tailored to HRC research (\cref{sec:hrc_model_set}).
Importantly, we do not focus on presenting a model set for benchmarking; rather, we aim at accelerating convergence toward a standardized task domain.
We have designed the model set to fulfill the following requirements (\cref{sec:features,sec:human_robot_interaction_design}):
\begin{inparaenum}[i)]
\item accessibility of the basic components and low cost of the final designs;
\item modularity, scalability and ease of re-use of the components;
\item applicability of the model set to a wide range of experiments in the broader HRC field;
\item scalability of task complexity;
\item lightweight design to suit the payload of the vast majority of collaborative platforms.
\end{inparaenum}
With this set of assembly tasks we also provide open-source software for robot perception and manipulation, in order to lower the barrier to entry for new users (\cref{sec:accompanying_software}).
By providing tools for perception, manipulation, and interaction we aim at promoting replicability; further, by encouraging consistency across experiments, we will add legitimacy and significance to contributions in the field.

\section{BACKGROUND AND RELATED WORK}
\label{sec:background}

Historically, while benchmarking proved crucial to achieve scientific, replicable research, its applicability to robotics has been limited due to the complexity of the field.
For this reason, previous work focused on establishing benchmarks in specific sub-fields of robotics research, e.g. manipulation~\cite{Calli2015}, motion planning~\cite{Cohen2012,Moll2015}, navigation~\cite{Madhavan2009,Sprunk2014}, service robotics~\cite{Iossifidis2005,Wisspeintner2009} or human-robot teamwork~\cite{Groom2007}.
Further, a number of robotic platforms have been designed with the specific purpose of mitigating the issue by fostering robotics research on shared hardware (e.g. PR2~\cite{PR2}, iCub~\cite{Metta2010}, and Poppy~\cite{Lapeyre2013}).
Closer to this work are instead robot competitions \cite{Iossifidis2005,AmazonPickingChallenge,DarpaRoboticsChallenge,Amigoni2013}, that propose a standard set of challenges to homogenize evaluation of robot performance.

\begin{table}
  \centering
  \caption{%
  Selection of previous work in HRC (in chronological order).
  First column: general domain; second column: approximate number of parts and tools involved in assembly tasks (in parenthesis, number of distinct parts when applicable); third column: nature of the actions performed by the robot.
  }

  \begin{tabular}{ l l c l }
    &   Domain &  Parts &  Robot actions \T \B \\
    \midrule
    \cite{deMello1990} & manufacturing & -- & pick, place \T\B\\
    \cite{Kosuge2000} & load sharing & -- & collaborative hold \T\B\\ %
    \cite{Breazeal2004,Hoffman2004} & buttons & 3 & press button \T\B\\ %
    \cite{Hinds2004} & warehouse & -- & pick, hold \T\B\\ %
    \cite{Kobayashi2005} & sweeping & -- & move \T\B\\ %
    \cite{Muller2006} & set table & -- & pick, place \T\B\\ %
    \cite{Beetz2008} & kitchen, set table & -- & -- \T\B\\ %
    \cite{Hoffman2007} & assembly & 10 & pick, pass \T\B\\ %
    \cite{Pardowitz2007} & kitchen, set table & -- & pick, place \T\B\\ %
    \cite{Dominey2009} & assembly & 5 (2) & hold, pass \T\B\\  %
    \cite{Cakmak2010} & virtual & -- & pick, place \T\B\\ %
    \cite{Shah2010} & assembly & 40 (20) & -- \T\B\\ %
    \cite{Shah2011} & assembly & -- & pick, place \T\B\\
    \cite{Tellex2011} & warehouse & -- & pick, place \T\B\\ %
    \cite{Wilcox2012} & manufacturing & -- & pick, place \T\B\\ %
    \cite{Dumora2012} & manufacturing & -- & collaborative hold \T\B\\ %
    \cite{Rozo2013} & assembly & 5 (2) & hold \T\B\\ %
    \cite{Knepper2013,Tellex2014a} & assembly & 5 (4) & pick, assemble \T\B\\ %
    \cite{Knepper2014} & assembly & 14 (6) & -- \T\B\\ %
    \cite{Hayes2014} & assembly & 8 & -- \T\B\\
    \cite{Hawkins2014} & assembly & 14 (3) & pick, place \T\B\\ %
    \cite{Vollmer2014} & assembly & 12 (6) & -- \T\B\\ %
    \cite{Gopalan2015} & diaper change & 2 & pick, pass \T\B\\ %
    \cite{Nikolaidis2015} & manufacturing & -- & hold \T\B\\ %
    \cite{Huang2015} & kitchen & -- & pick, pass \T\B\\ %
    \cite{Hayes2016} & assembly & 3 & -- \T\B\\ %
    \cite{Toussaint2016} & assembly & 5 & pick, pass, hold \T\B\\ %
    \cite{Roncone2017,Mangin2018} & assembly & 4 (4) & pick, place, pass, hold \T\B\\ %

  \end{tabular}

  \label{tab:previous-work}
\end{table}

A great variety of tasks has been used in human-robot collaborative studies, each featuring specific properties that address one or several aspects of the collaboration.
It is therefore particularly challenging to provide tools to standardize research in the field.
A selection of relevant work is shown in \cref{tab:previous-work}.
We included papers that either explicitly or implicitly target HRC applications, with particular focus on contributions that feature practical deployments or concrete task domains.
This selection criteria resulted in a wide variety of final applications, with works that e.g. focused on evaluating the communication between peers~\cite{Breazeal2004,Hoffman2004,Dominey2009,Tellex2011,Roncone2017}, leveraging the collaborative domain to target physical human-robot interactions~\cite{Kosuge2000,Dumora2012}, or proposing a teaching scenario where the robot learns a task model~\cite{Cakmak2010,Hayes2014,Toussaint2016}.
This broad landscape of applications is then summarized in \cref{tab:previous-work} in terms of the type of actions implemented on the robot platform. Remarkably, there is consistent overlap in robot capabilities; they are typically reduced to pick and place movements, button presses, hold actions, object passing, or basic assemblies.
When applicable, the table also shows a basic measure of task complexity in terms of number of parts that need to be assembled. This directly relates to the intricacy of the task that needs to be represented (and eventually learned) by the robot, and the complexity of the physical interaction with the human partner.
Of particular importance is the fact that these tasks are often tailored to the experiment, robot, and research space involved, and do not allow for a high degree of replicability.
Further, there is frequently a hazy distinction between what constitutes the experimental scenario and the theoretical contribution itself.
For example, papers addressing kitchen assistance domains such as setting the table feature a set of robot skills very similar to contributions that target advanced manufacturing.
To alleviate these issues, we propose a framework aimed at converging toward a common language for human-robot collaboration.
It is meant to provide a set of assembly tasks from which researchers can build a variety of HRC applications.
Importantly, we demonstrate how the proposed model set covers the majority of the works detailed in \cref{tab:previous-work} in terms of both task complexity and robot capabilities.
In addition to that, we provide a set of software tools and libraries to facilitate the development of such experiments.

\section{THE HRC MODEL SET} %
\label{sec:hrc_model_set}

We propose a model set for human-robot collaboration experiments, aimed at standardizing repeatability and accelerating research in the field.
We address the fact that, despite HRC being a growing area of study within the broader robotics umbrella, the difficulty to reproduce contributions in the field hinders its progress.
To this end, we aim at providing a cohesive and comprehensive set of building blocks with the goal of: i) covering the needs of typical HRC works in terms of complexity and robot skills; ii) facilitating the setup of collaborative experiments to accelerate research in the field.
In the following, we detail the hardware components that constitute the model set (\cref{sub:hardware_components}), and we later present a number of prototypical designs built on top of said components (\cref{sub:example_designs}).

\subsection{Hardware Components} %
\label{sub:hardware_components}

{%

\begin{table*}
  \centering
  \caption{List of components for the example designs detailed in \cref{sub:example_designs}, with number of parts, estimated costs and weights.}

  \begin{tabular} { r c c c c c c c c c c c c c c c c c c}
  & \multicolumn{8}{c}{\textsl{Brackets}} & \textsl{Wooden} & \multicolumn{3}{c}{\textsl{Plywood} [in]} & \multicolumn{2}{c}{\textsl{Screws} [in]} & Number & Cost & Weight \T \B \\
  \cmidrule(r){2-9} \cmidrule(r){11-13} \cmidrule(r){14-15}
          & CB & CL & CR & F &  S90 &  S180 &  T180  &  T &  \textsl{Dowels}  &  6x8 &  6.75x2.5 &  6x16 &  1/2 &  1/4 & of Parts &  [\$] & [g] \T \B \\
  \midrule
  \textsl{Table}   & 0  & 0  & 0  & 4  &  0  & 0 & 0 & 4 &  4 & 1 & 0 & 0 &  8 &  8 &  29 &  58 &  208 \T \B  \\
  \textsl{Chair}   & 2  & 1  & 1  & 4  &  0  & 0 & 0 & 2 &  7 & 1 & 1 & 0 & 14 &  8 &  41 &  84 &  312 \T \B  \\
  \textsl{Shelf}   & 0  & 0  & 0  & 4  &  4  & 0 & 0 & 4 &  8 & 2 & 0 & 0 & 16 & 16 &  54 & 103 &  398 \T \B \\
  \textsl{Console} & 0  & 0  & 0  & 8  & 12  & 4 & 4 & 4 & 24 & 6 & 0 & 2 & 48 & 24 & 136 & 302 & 1299 \T \B  \\

  \end{tabular}
  \label{tab:modelset:list}
\end{table*}
}

The components of the model set can be divided in four main groups: \textsl{screws}, \textsl{dowels}, \textsl{plywood}, and \textsl{brackets}. These basic constituents can be assembled in a variety of different configurations, as detailed in \cref{sub:example_designs} (please refer to \cref{tab:modelset:list} for a comprehensive list of all the parts that belong to the model set).
As mentioned in \cref{sec:introduction}, we aim at providing a set of accessible tools \textsl{by design}.
For this reason, the screws, dowels, and plywood can be easily purchased at a hardware store, craft supply, or online. For the purposes of this work, we use dowels with a diameter of $\unit[\nicefrac{1}{2}]{in}$, standard \#$4$ screws in $\unit[\nicefrac{1}{2}]{in}$ and $\unit[\nicefrac{1}{4}]{in}$ lengths, and birch plywood with a thickness of $\unit[\nicefrac{1}{8}]{in}$ that has been laser cut into rectangles of different sizes ($\unit[6\times8]{in}$, $\unit[6.75\times2.5]{in}$, and $\unit[6\times16]{in}$ respectively).
The only components that have been custom designed to suit the purposes of this work are the brackets, which are meant to connect dowels and plywood panels in a variety of ways, and are held in place by the screws. Their CAD models have been released under a \textsl{CC-by-sa} open-source license, and are freely available on GitHub\footnote{ \href{https://github.com/ScazLab/HRC-model-set}{\texttt{github.com/scazlab/HRC-model-set}} hosts both STL and SolidWorks\textsuperscript{\textregistered} files for the brackets, along with specifications for 3D printing, tutorials for assembling the example designs in \cref{sub:example_designs}, and reference links for purchasing dowels, plywood panels and screws.}.

\begin{figure}
  \centering
  \subfloat[`\textsl{Top}' (T)]{\includegraphics[width=.49\linewidth]{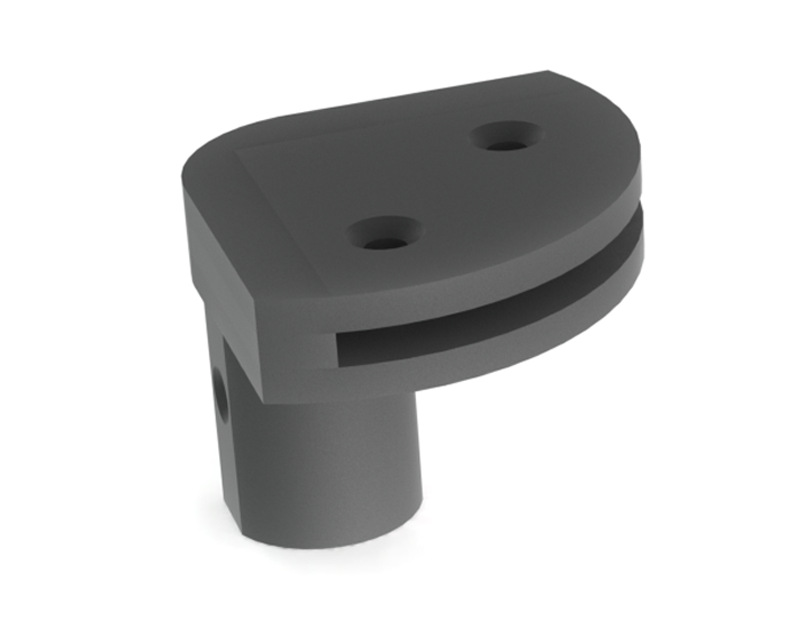}\ \label{fig:bracket:T}}
  \subfloat[`\textsl{ShelfBracket $180^{\circ}$}' (S180)]{\includegraphics[width=.49\linewidth]{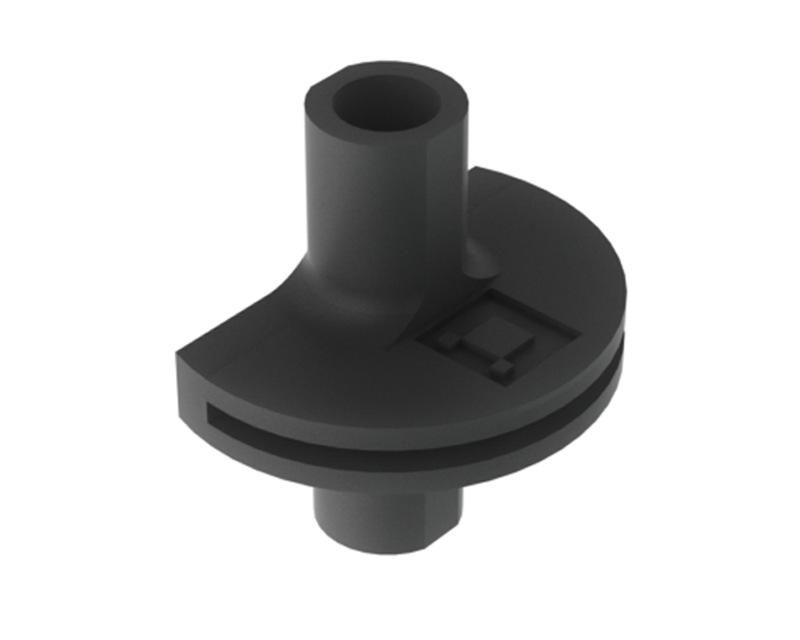}  \label{fig:bracket:S180}}

  \caption{Example 3D renderings for some of the brackets available in the model set. The design of the bracket was carried out with modularity in mind. From the basic `\textsl{Top}' element in \cref{fig:bracket:T}, eight derivative brackets have been developed in order to extend the type of final designs the model set allows for. Optionally, a fiducial marker can be embedded in the design, to facilitate perception (\cref{fig:bracket:S180}).}\label{fig:bracket}
\end{figure}

In order to be able to flexibly extend the capabilities of the model set, we started from the basic element depicted in \cref{fig:bracket:T}, which simply connects a dowel with a plywood panel.
This bracket, referred to as `\textsl{Top}' (T) in \cref{tab:modelset:list}, is particularly suited for connecting e.g. a leg with the top of a table.
Importantly, minimal changes to its design lead to new connections that can scale according to the specific needs of the user. In this work, we have designed a set of seven additional modules, hereinafter referred to as `\textsl{ChairBack}' (CB), `\textsl{ChairBracketL}' (CL), `\textsl{ChairBracketR}' (CR), `\textsl{Foot}' (F), `\textsl{ShelfBracket $90^{\circ}$}' (S90), `\textsl{ShelfBracket $180^{\circ}$}' (S180, \cref{fig:bracket:S180}), and `\textsl{Top $180^{\circ}$}' (T180). Each of these modules is used as a component of the prototypical assemblies described in \cref{sub:example_designs} (whose constituent parts are listed in \cref{tab:modelset:list}).

\subsection{Example Designs} %
\label{sub:example_designs}

\begin{figure}
  \centering
  \subfloat[  Table]{\includegraphics[height=.35\linewidth]{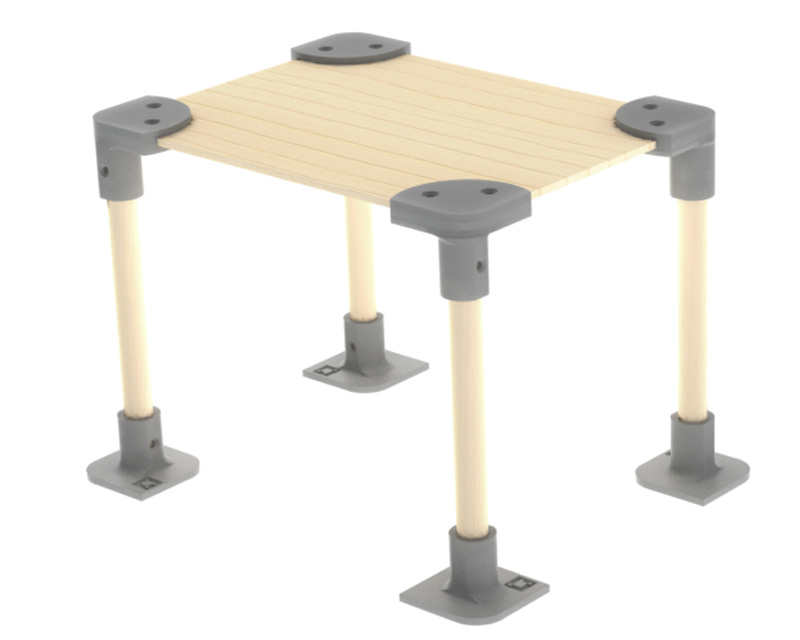}\quad }
  \subfloat[  Chair]{\includegraphics[height=.35\linewidth]{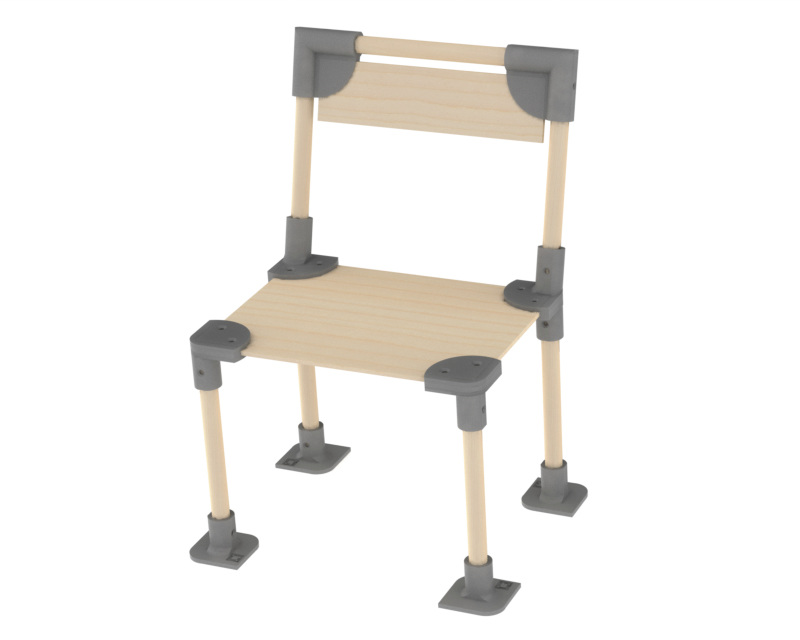} }\\
  \subfloat[  Shelf]{\includegraphics[height=.35\linewidth]{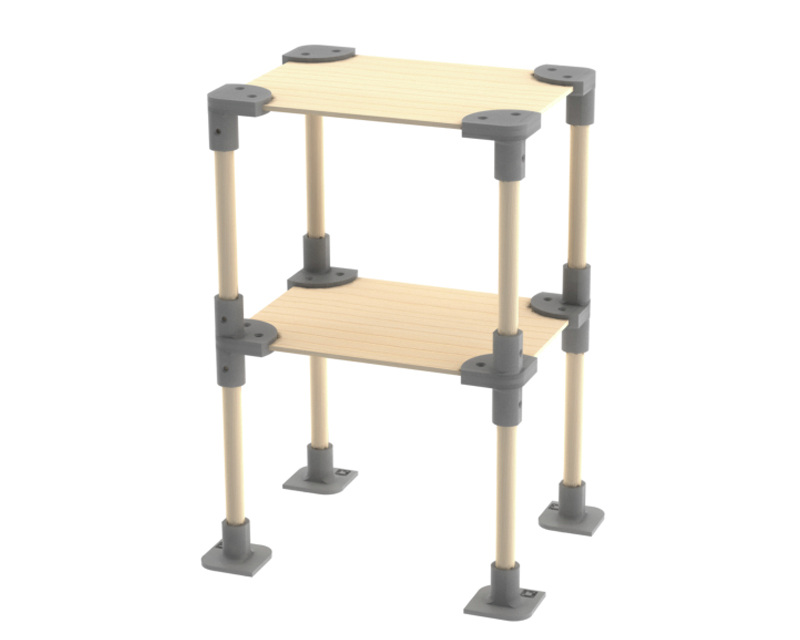}\quad }
  \subfloat[Console]{\includegraphics[height=.35\linewidth]{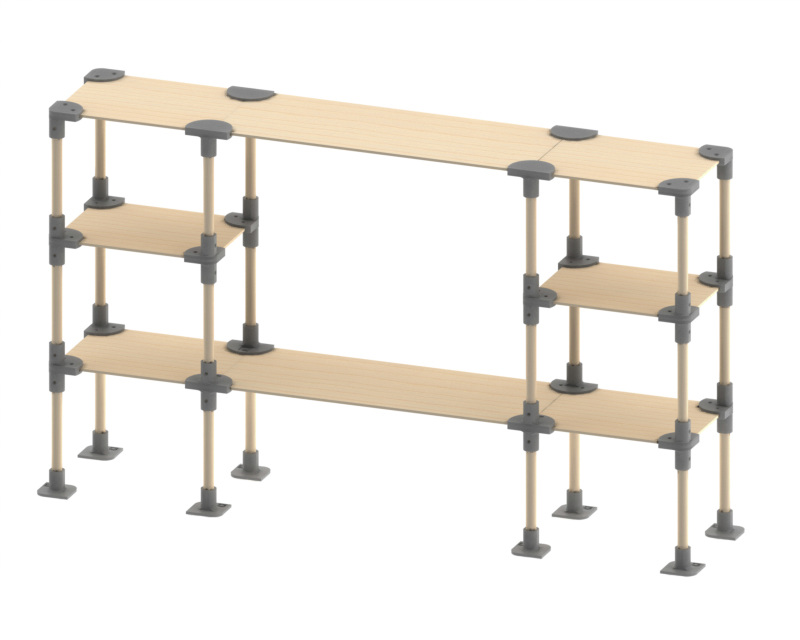}  }

  \caption{3D renderings of the example designs detailed in \cref{sub:example_designs}. Please refer to \cref{tab:modelset:list} for a list of the components needed for each design.}\label{fig:example_designs}
\end{figure}

\begin{figure}
  \centering
  \subfloat[Table]{\includegraphics[height=.44\linewidth]{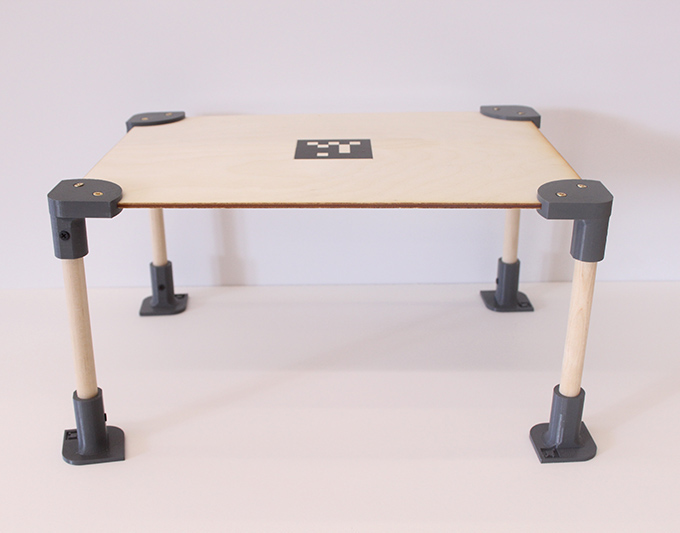}\ }
  \subfloat[Chair]{\includegraphics[height=.44\linewidth]{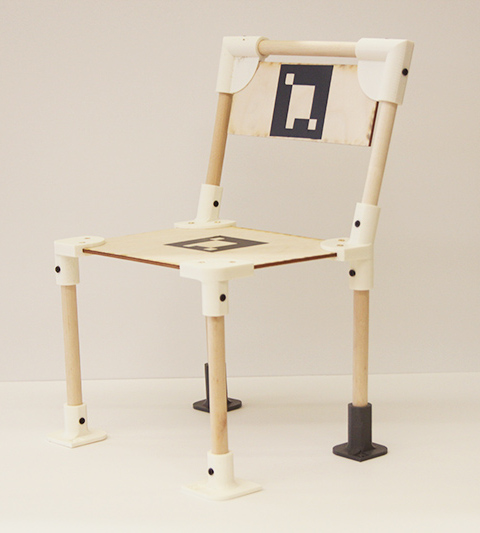}  }

  \caption{Final constructed objects (see \cref{sub:example_designs}). Similarly to brackets (\cref{fig:bracket}), flat surfaces such as the plywood panels can be equipped with adhesive fiducial markers to ease perception.}\label{fig:final_designs}
\end{figure}

As detailed in the previous Sections, the model set has been designed with modularity and scalability in mind. The main idea is to provide a number of basic parts that can be used to create a variety of different objects, tailoring to the specific needs of the user and the platform used.
For the purposes of this work, we present four prototypical objects built from the parts described in \cref{sub:hardware_components}: a table, a chair, a shelf and an entertainment console. \cref{fig:example_designs} illustrates a set of 3D renderings for the four basic configurations, whereas \cref{fig:final_designs} shows the final objects for table and chair.

These four designs have been devised to showcase the flexibility of the model set. As highlighted in \cref{tab:modelset:list}, they differ in terms of number of parts, cost, and weight.
The simplest construct---the table---is composed of only $29$ parts, whereas the console represents the other extreme, with $136$ parts---of which $72$ are screws which need to be screwed in to complete the assembly.
Further, some configurations feature a higher level of modularity than others:
for example, combinations of $10$ unique parts can be made into either the chair ($41$ total parts) or the more complex console ($136$ parts).
Either may prove useful for specific experimental requirements.
This ability to support a number of final designs with different properties enables granular control of the interaction between the robot and the human partner, and is considered an important asset of the model set.
Please refer to \cref{sec:features} for a detailed discussion on the features of the example designs in particular, and the model set at large.

\section{FEATURES AND USAGE}
\label{sec:features}

In this section, we detail the peculiar features of the proposed model set, by focusing on how it is easy to distribute, compatible with many robotic platforms, and an efficient solution for a variety of collaborative experiments.

\subsection{Design Considerations for Robot Interactions} %
\label{sub:design_considerations_for_robot_interactions}

It is common practice in robotics research to shape the environment to suit the physical embodiment of the robot and to conform with the constraints dictated by the experiment.
Similarly, an important requirement for the design of the HRC model set was for it to be easily deployed to a variety of robotics platforms and collaborative setups.
To this end, a number of considerations were taken into account. In the following, we highlight the design choices behind the HRC model set to ease perception, manipulation and robot control.

Although a wide range of state of the art algorithms can be exploited in HRC experiments, we designed the model set to comply with minimal requirements.
For what concerns the perception problem, our only prerequisite is the availability of standard RGB cameras.
As detailed in \cref{sec:accompanying_software}, we provide a marker-based software to detect flat parts such as the plywood panels, if conveniently equipped with proper markers (cf. \cref{fig:example_designs}).
For dowels, we have created a marker made up of a flat piece of plywood and a cable clip that attaches to them (see \cref{fig:camera_views:left}).
Lastly, brackets can be printed with a unique marker identifier that takes advantage of the same technique, as shown in \cref{fig:bracket:S180}.

The design of the model set also takes into consideration standard hardware in terms of manipulation and grasping.
With the only exception of the screws---that are too small for most robotic platforms to date---it is possible to individually pick up the totality of the parts that compose the model set.
Flat surfaces are particularly suitable for vacuum grippers; if a vacuum gripper is not available, they can be vertically positioned in order to facilitate pickup by a standard parallel gripper.
Dowels and brackets offer the same type of affordance to a robot, as they are suitable for e.g. parallel grippers that can grasp by aligning with the longitudinal axis.
Furthermore, it is always possible to conveniently locate smaller objects such as screws into bigger containers that are easier to manipulate, such as the boxes used in \cref{fig:camera_views:right}.

Lastly, to overcome the limitations of standard collaborative robots' payloads and ease experiment design, we favored a scaled down furniture design over e.g. commercially available solutions.
We employed two strategies: i) reduce the size of each final design to only one fourth of what it would be as a real furniture item; ii) use lightweight components.
This latter aspect led to the choice of plywood among a variety of possible materials, and weight considerations when designing the brackets.
As a consequence of this, the heaviest bracket weighs only $\unit[18]{g}$ and the heaviest plywood panel used in the assembly is $\unit[127]{g}$. Importantly, even the entertainment console has a still reasonable weight of $\unit[1.3]{kg}$ in total (cf. \cref{tab:modelset:list})---well below the payload limit of most robot platforms.

\subsection{Impact of 3D Printing on Dissemination} %
\label{sub:impact_of_3d_printing_on_dissemination}

3D printing is a quickly evolving technology that facilitates distribution of design; if compared to common manufacturing methods, it allows to contain costs when small batches and multiple iterations are required~\cite{Ma2013,Lapeyre2013}.
Both these features are essential to support the level of dissemination that the HRC model set targets.
As mentioned in \cref{sub:hardware_components}, we open-sourced the design of all the brackets, with the goal of encouraging external contributions. They are available online alongside detailed instructions on how to purchase all the components of the model set, and how to assemble the prototypical configurations.
Importantly, the brackets can be printed with any commercially available 3D printer. In order to guarantee that plywood and dowels snap into brackets without being too loose, an additive technique that employs soluble material and a layer height of $\unit[0.15]{mm}$ are recommended.
For the purposes of this work, we employed a fused deposition modeling technique with ABS+ plastic. %
While the first solution guarantees clean inner edges on the inside of the brackets, ABS+ plastic can withstand multiple utilizations and structurally support the weight of each final configuration.

\subsection{Scalability} %
\label{sub:scalability}

An important asset of the model set is that its modular nature allows for reproducibility, reuse, and scalability.
For instance, the wooden parts, in particular brackets and dowels, can be reused in numerous assemblies of varying complexity.
Indeed, all the prototypical designs described in \cref{tab:modelset:list} are characterized by a high degree of reusability: each assembly shares at least $60\%$ of its components with other objects. For simpler designs such as the table, the overlap increases to $100\%$, which means that the table can be constructed entirely from parts used to create a chair or a shelf.
This overlap in terms of components allows to deploy a variety of construction tasks---and hence experimental scenarios---without the need to purchase additional hardware.
This enables to explicitly treat the complexity of the assembly (for example in terms of number of distinct parts) as an experimental variable, that can be tuned by combining the same components into different pieces of furniture.
Finally, distributing the 3D models of the brackets enables third parties to adapt the furniture designs to specific research spaces and robot capabilities, or to extend the model set to investigate novel scenarios.
For example, in order to create a full-scale chair, one can print the brackets at $4$ times their size, use plywood that is $\unit[\nicefrac{1}{2}]{in}$ in thickness, dowels that are $\unit[2]{in}$ in diameter, and \#$10$ screws in $\unit[1]{in}$ and $\unit[2]{in}$ lengths.

\subsection{Cost Considerations} %
\label{sub:cost_considerations}

Accessibility of the model set is guaranteed by keeping costs reasonably low. As mentioned in \cref{sub:hardware_components}, most of the parts that compose the model set can be easily retrieved online or at hardware stores for a small price.
For the custom designed brackets, we provide an estimate of their cost using the SolidWorks\textsuperscript{\textregistered} cost evaluation for 3D printing, which includes the cost of outsourcing the printing if an appropriate device is not available.
Results are shown in \cref{tab:modelset:list}: for what concerns the majority of the proposed configurations (table, chair, and shelf), it is possible to purchase the totality of the hardware from under \$$100$; the only outlier being the entertainment console, which is the extreme case composed of $136$ different parts, and has a cost of \$$300$.

\section{HUMAN--ROBOT INTERACTION DESIGN} %
\label{sec:human_robot_interaction_design}

The model set allows for a series of assembly tasks to be performed in collaboration between a human and a robot.
Particular attention has been put into providing multiple dimensions of variability in the design of experiments that can cover the broad range of applications HRC research.
Not only does it enable a large spectrum of experiments, but it explicitly exposes \emph{experimental variables} for finer evaluation of the algorithms.
Firstly, in order to be as close as possible to real collaborative environments, we impose the requirement of needing at least one tool for the assembly.
As detailed in \cref{sub:hardware_components}, only one type of screwdriver is needed by default since the screws are all of the same type, but the user is free to adapt the design to allow for multiple tools.
In order to succeed in the construction of each of the objects detailed in \cref{sub:example_designs}, the following is a list of the necessary actions:
\begin{inparaenum}[i)]
  \item pre-drilling screw holes into plywood panels;
  \item retrieving components;
  \item retrieving tools;
  \item mounting parts together;
  \item holding parts in place to support screwing;
  \item screwing parts;
  \item cleaning up the workplace.
\end{inparaenum}
An important asset of the proposed model set is that the combination of skills required for each task---screwing, holding, bringing parts---integrates well with the different capabilities of robot and human.
This allows for easy incorporation of robot participation and support into a task that would be difficult for a human to complete alone. For example, human participants might find it particularly difficult to screw a bracket onto a dowel without the support of the robot, who can help by holding the dowel in place (cf. \cref{fig:setup}).
That is, a potential experiment can force the two partners to a certain level of interaction \emph{by design}.

Further, it is possible to tune role assignment and task allocation between peers, if needed.
This is a direct consequence of the design choices that led to the proposed model set (cf. \cref{sub:design_considerations_for_robot_interactions}), which is composed of a series of parts that a standard robotic platform can interact with.
That is, although some tasks pertain to the human exclusively because of their superior manipulation and perception skills, there exists some degree of overlap in terms of actions that can be performed by both agents.
Consequently, it is possible to devise an experiment in which there is a clear compartmentalization of the roles the two partners have.
Similarly to \cite{Hawkins2014,Gopalan2015,Roncone2017}, specific actions (e.g. bringing components) can be assigned to the robot, whereas the human can be entrusted to perform only the actions that require fine manipulation skills (e.g. screwing).
Conversely, the respective areas of competence can be artificially mixed and the two agents can be forced to compete for the same resource \cite{Shah2011}.

Lastly, the modularity of the model set allows granular control on the amount of information available to the human and the robot.
One of the issues the majority of works in the literature face is that, as of now, the human is far superior to the robot in terms of both perception and manipulation skills.
This dissymmetry of skills and available information between the two partners fundamentally limits the amount of possible interactions allowed in standard HRC experiments.
Without diverging from the scope of this work, it is possible to customize the model set in order to balance out the information asymmetry, and consequently favor the robot.
For example, brackets can be customized to be compatible with multiple plywood depths and dowel widths, and their detection can become particularly challenging for a human participant.
Conversely, this granular information can be implicitly embedded in the perception system of the robot---via e.g. the fiducial markers (\cref{sec:accompanying_software}).
That is, it becomes easier for the robot to pick up the correct dowel for a specific bracket, increasing the efficacy and effectiveness of the robot platform in the task domain.
Similarly, it is also possible to obfuscate the task goal by for example providing a set of components that are not needed for the assembly and neither partners (or only one of them) know how to use. These use cases are of particular interest for the study of collaboration with ambiguity in the task structure.
As mentioned, the modular structure of the furniture set also makes task complexity an experimental variable that allows the experimenter to modulate the total number of parts involved in the final construction.

\paragraph*{Evaluation of human-robot collaborations}

Standardization of human-robot collaborative interactions is currently lacking in the field.
Although benchmarking is out of the scope of this work, the proposed model set constitutes a first step toward achieving a common ground for repeatable, reusable algorithms.
In related work \cite{Roncone2017,Mangin2018}, examples evaluations in HRC settings are presented; they provide insights on how to set the baseline on the collaborative efficiency of the proposed algorithms, as well as how to compare these algorithms across several conditions.
In particular, two metrics are introduced:
 i) the ratio of \textsl{valid actions} and \textsl{robot errors} as a criteria for robot performance---useful to evaluate the robot w.r.t. a canonical or reference behavior;
ii) the \textit{completion time} of the overall assembly to provide an objective evaluation of the interaction.
Additionally, one can also consider the human perception of the robot (e.g. through questionnaires) or evaluate the cognitive load on the human.

\section{ACCOMPANYING SOFTWARE}
\label{sec:accompanying_software}

Our model set is intended as an inexpensive, accessible, reproducible, and scalable resource for easier development of HRC experiments.
In conjunction with this reference set, we have also developed a complementary collection of software tools and libraries compatible with the Robot Operating System (ROS, \cite{ROS2009}), the most used middleware to date for robotics research and industry applications.
The accompanying software has been released under the \textsl{LGPLv2.1} open-source license, and it is freely available on GitHub\footnote{ \href{https://github.com/scazlab/human_robot_collaboration}{\texttt{github.com/scazlab/human\_robot\_collaboration}} hosts both the source code for the software library, and some example applications.}.
Its main purpose is to facilitate prospective users with the adoption of the model set in collaborative experiments.
As such, the use of the software libraries is advantageous in contexts that are similar to what is generally envisioned by the authors.
For the purposes of this work, we have developed: i) a collection of robot-independent perception tools tailored to the constituent parts of the model set; ii) a robot controller that exposes a set of high-level actions to the user.
Collectively, they target a generic human-robot collaborative setup, with the aim of equipping a standard robot with enough skills for it to effectively locate, manipulate, and interact with parts and tools.
It is worth noting how the software presented in this work has been implemented on a Baxter Research Robot, a widely used platform for research in HRC (cf. \cref{fig:setup}). Nonetheless, it is easy to customize to any robotic platform that shares similar hardware features and is provided with a kinematic model in URDF form.

\begin{figure}
  \centering
  \subfloat[Dowels and plywood panels.]{\includegraphics[height=.38\linewidth] {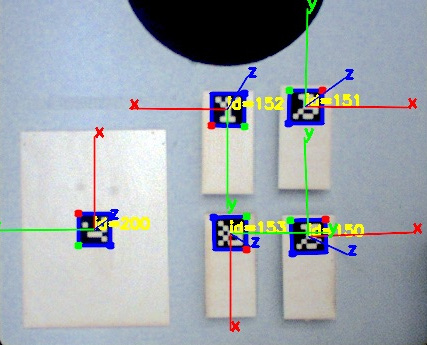}\label{fig:camera_views:left} \ }
  \subfloat[Screwdriver, brackets and screws.]{\includegraphics[height=.38\linewidth]{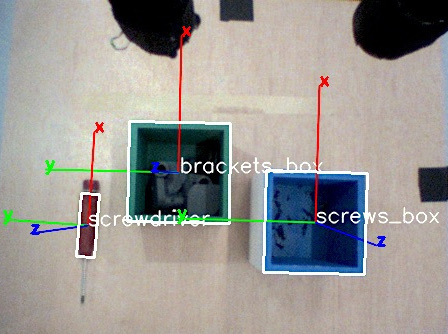}\label{fig:camera_views:right}  }

  \caption{Snapshots of the left and right camera views from the Baxter's end effectors, that show the fiducial marker algorithm (\cref{fig:camera_views:left}) and the HSV-based detection system (\cref{fig:camera_views:right}) in action.}\label{fig:camera_views}
\end{figure}

As mentioned in \cref{sub:design_considerations_for_robot_interactions}, a system based on fiducial markers is provided to support precise 6D localization of the majority of components of the model set.
Such markers can be directly engraved in the brackets during printing, or taped to the plywood panels. For dowels, we provide a marker support that clips to them.
The perception system is based on ARuco \cite{Aruco2014}, an OpenCV-based library able to generate and detect fiducial markers; it is depicted in \cref{fig:camera_views}.
Additionally, the robot should also be able to interact with smaller objects and tools---and in general objects that are difficult to tag with a marker.
To this end, we developed an additional HSV-based color segmentation algorithm, that provides similar 6D localization features as the ARuco based system.
Importantly, the HSV software requires minimal information to operate---namely the color of the objects to track and their physical sizes in the real world.
As shown in \cref{fig:camera_views:right}, the proposed algorithm is capable of successfully identifying and locating a tool (the screwdriver), and containers that store smaller parts such as brackets and screws.
Provided a robot-camera calibration system is available, by coupling the feedback from either the ARuco-based system or the HSV-detection algorithm with the robot's kinematic model it is then possible to track objects into the 3D operational space of any robotic platform.
Importantly, both perception systems run at the standard frequency for robot camera systems, that is $30$ Hz. This proves useful in order to achieve precise localization and grasping, in particular if the camera is positioned on the end-effector.
In such settings, it is possible to implement visual servoing with a feedback loop that is close to the interaction itself, i.e. with an estimation that continuously improves as the robot gets closer to the target. Please refer to the accompanying video (also available at \href{https://youtu.be/09Zflg7ZzKU}{https://youtu.be/09Zflg7ZzKU}) for an example of the robot interacting online with an human participant.

On top of the perception algorithms highlighted above, we implement a control library for high-level actions. The goal is to provide a set of basic capabilities for the robot to interact with the components, thus lowering the barrier to entry for new users of the model set.
Similarly to the approach used for the perception software, our goal is to provide a set of robot skills that use minimal equipment and do not exploit complex machinery, such as room-scale RGB-D sensing or 3D tracking.
For this reason, we implemented the control software on the Baxter robot, using the default hardware and software SDK (see \cref{fig:setup}).
In the context of this work, we developed two state-less position controllers (one for each of the robot's arms), able to interface with the user through a library of high-level actions (in the form of ROS services).
These actions can range from single-arm commands---such as `\textsl{pick screwdriver}'---to complex bi-manual trajectories---e.g. `\textsl{handover dowel from left to right arm}'---or instructions that require physical interaction with the human---such as `\textsl{hold plywood panel}' (to support screwing).

Lastly, a number of context-based, multi-modal communication channels are provided to support human-robot interactions.
The exposed layers are:
\begin{inparaenum}[i)]
\item a web interface for high-level action management;
\item a feedback channel on the Baxter's head display;
\item a text-to-speech and a speech-to-text system for voice interactions;
\item a physical interaction layer based on force feedback;
\item an error/emergency channel to send error messages to the robot during operation.
\end{inparaenum}
It is worth noting how the utilization of these interaction layers is at the discretion of the experimenter and is contingent on the specific requirements that each experiment entails; nonetheless, they are deemed a relevant component aimed at supporting HRC experiments and complementing the basic perception and control skill-set of the robot.
The model set (and its accompanying software) has been already adopted in related work \cite{Roncone2017,Mangin2018}.

\section{DISCUSSION}

In this paper, we introduce a model set for furniture assembly tasks that is intended to bring human-robot collaboration research closer to standards for replicability.
We provide references and 3D models of the parts composing the set, as well as open-source software for robot perception and human interaction.
In particular, we chose to implement the task set as an assembly task, which is close to collaborative manufacturing domains.
By analyzing relevant research in the literature, we substantiate our choice by demonstrating how the model set covers, the majority of task domains and robot actions presented in previous work.
We expect our contribution to simplify the design and deployment of future experiments in the field, and to form a language around which to establish reference tasks for HRC research.

The ability to accurately assess and compare contributions to the field is indeed essential for further advancing HRC research. With this in mind, we intend for the proposed model set to foster the development of greater standardization.
In particular, we provide examples of how to evaluate collaboration performance and expect future works to leverage our contribution to converge toward benchmarks for collaborative domains, and most notably to
establish reference measures of HRC task complexity.
Importantly, the range of assembly tasks conceivable by the set will not necessarily cover all the potential ramifications of the field.
Nevertheless, we are confident that it will prove useful to the rapid, cost-effective development of future experiments.
Moreover, we expect future works to identify unforeseen pitfalls and either extend the set or identify new complementary task domains.
Finally, a noteworthy direction for future work is to integrate and compare independent contributions under the same experimental setup.
Such efforts will enable thorough assessment of their interoperability, scalability, potential for applications, and eventual limitations.
 
\addtolength{\textheight}{-2cm}   %

\bibliographystyle{IEEEtran}
\bibliography{root}

\end{document}